\documentclass[manuscript,screen,review=false]{acmart}
\usepackage{enumitem,multicol}
\usepackage{wrapfig}
\AtBeginDocument{%
  \providecommand\BibTeX{{%
    \normalfont B\kern-0.5em{\scshape i\kern-0.25em b}\kern-0.8em\TeX}}}

\setcopyright{acmlicensed}
\copyrightyear{2024}
\acmYear{2024}
\acmDOI{}
\acmConference[HRI]{Make sure to enter the correct
  conference title from your rights confirmation email}{ March 11 -- 15, 2024}{Boulder, CO}
\acmBooktitle{Workshop on HRI Education at HRI '24: ACM/IEEE International
Conference on Human Robot Interaction (HRI),
 March 11 -- 15, 2024, Boulder, CO} 
\acmISBN{978-1-4503-XXXX-X/18/06}

\begin{document}

\title{Hey, Teacher, (Don't) Leave Those Kids Alone: Standardizing HRI Education}

\author{Alexis E. Block}
\email{alexis.block@case.edu}
\orcid{0000-0001-9841-0769}
\affiliation{%
  \institution{Case Western Reserve University}
  \streetaddress{P.O. Box 1212}
  \city{Cleveland}
  \state{Ohio}
  \country{USA}
  \postcode{43017-6221}
}

\renewcommand{\shortauthors}{Block}

\begin{abstract}
Creating a standardized introduction course becomes more critical as the field of human-robot interaction (HRI) becomes more established. This paper outlines the key components necessary to provide an undergraduate with a sufficient foundational understanding of the interdisciplinary nature of this field and provides proposed course content. It emphasizes the importance of creating a course with theoretical and experimental components to accommodate all different learning preferences. This manuscript also advocates creating or adopting a universal platform to standardize the hands-on component of introductory HRI courses, regardless of university funding or size. Next, it recommends formal training in how to read scientific articles and staying up-to-date with the latest relevant papers. Finally, it provides detailed lecture content and project milestones for a 15-week semester. By creating a standardized course, researchers can ensure consistency and quality are maintained across institutions, which will help students as well as industrial and academic employers understand what foundational knowledge is expected.

\end{abstract}

\received{21 January 2024}

\maketitle
\vspace{-0.5cm}
\section{Introduction}
\label{sec:intro}
While many excellent human-robot interaction (HRI) researchers teach some form of a course on the topic, a standardized introductory course still needs to be created. Many researchers offer graduate-level versions of the course, which may focus more on the specific researcher's focus area of HRI or be a seminar-style class with reading weekly scientific articles but often feature little to no lecturing on fundamental topics. At the end of an introductory human-robot interaction course, I believe every student should have the tools to:
\vspace{-0.15cm}
\begin{itemize}
    \item \textbf{Read}, \textbf{understand}, and \textbf{discuss} recent \textbf{literature} 
    \item Have a \textbf{comprehensive overview} of the entire field of \textbf{HRI} (not just a subset)
    \item \textbf{Design a user study} with human participants 
    \item \textbf{Develop} a \textbf{hands-on interaction} with a \textbf{real robot}
    \item \textbf{Analyze} and \textbf{evaluate} experimental \textbf{data}
    \item \textbf{Communicate} their \textbf{findings}
\end{itemize}
\vspace{-0.15cm}
This manuscript describes the importance of an introductory course containing theoretical and experimental components in Section~\ref{sec:theo_exp}, while Section~\ref{sec:platform} advocates for adopting or creating a universal robotic platform that all introductory students will gain experience with through a semester-long project, regardless of university funding or size. I also see significant value in some teaching technical papers at the undergraduate level, and I recommend a comprehensive way to do this in  Section~\ref{sec:papers}. Section~\ref{sec:content} more explicitly outlines the proposed course content for such an introductory course based on my experience designing and teaching such a course in Fall 2023. Finally, Section~\ref{sec:conc} summarizes this short paper emphasizing supporting students throughout the learning process.

\begin{wrapfigure}{r}{0.4\textwidth}
    % \vspace{-0.5cm}
    \includegraphics[width=0.38\textwidth]{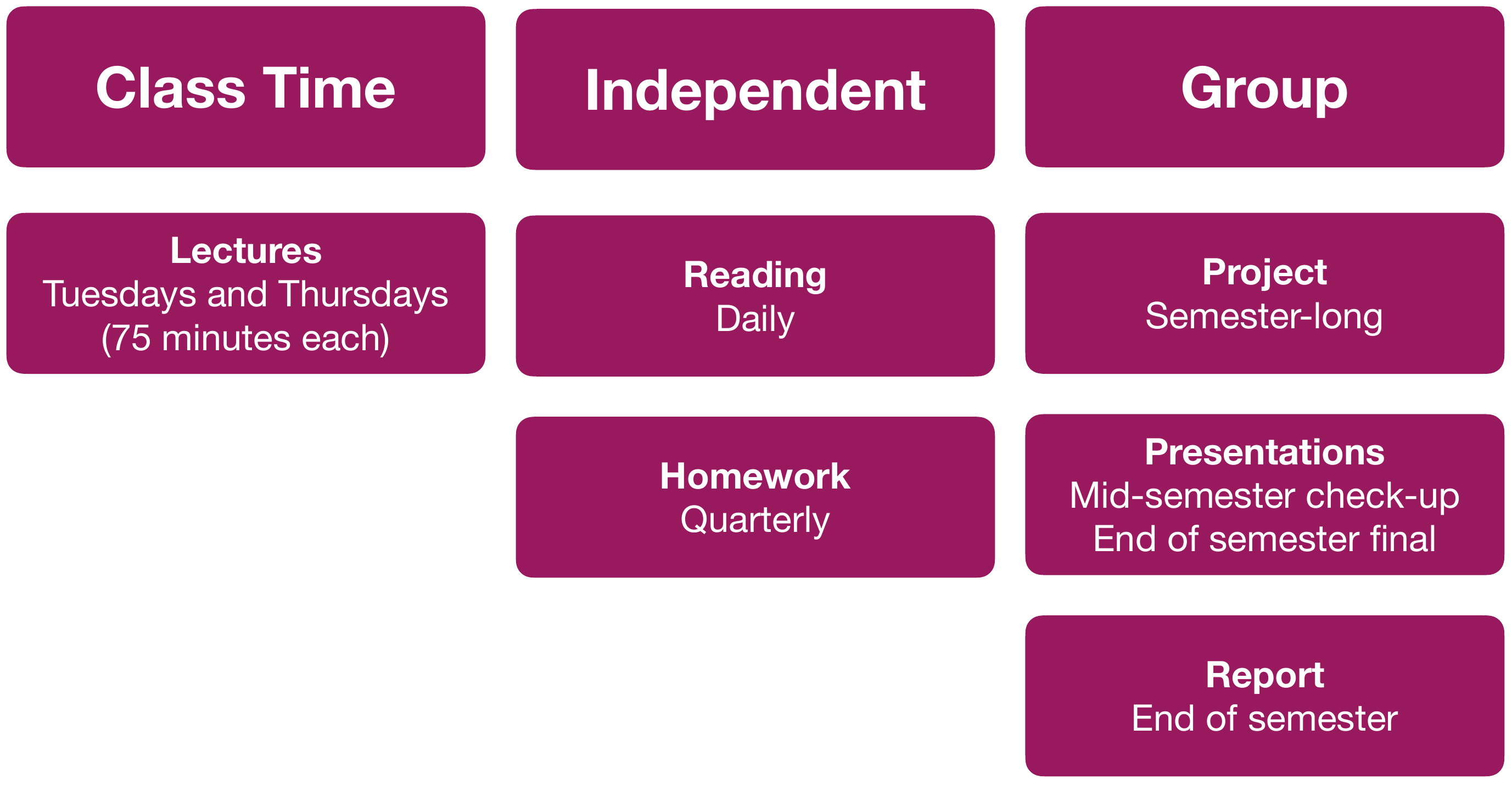}
    \vspace{-0.3cm}
    \caption{The class structure for the author's introductory HRI course. Course content was broken up between class time, independent, and group activities.}
        % \caption{The suggested topics for an introductory HRI course. The two blocks on the left of the figure represent topics that take place throughout the semester, the two columns on the right should be followed in chronological order.}
    \label{fig:overview}
\end{wrapfigure}

\vspace{-0.25cm}
\section{The Case for Theoretical AND Experimental Education}
\label{sec:theo_exp}

The familiarity of lectures provides undergraduate students with the necessary and critical new content of an unfamiliar subject in a digestible format. However, research shows that student retention of content increases by combining and increasing the number of learning methods utilized in the class activity \cite{stice1987using}. Thus, by incorporating multiple learning activities, like lectures, readings, discussions, hands-on assignments, and projects with demonstrations, an introductory HRI course can accommodate various learning preferences \cite{pashler2008learning} and encourage active participation, improving content retention. Finally, research in robotics education has shown that hands-on experimental components increase motivation, creativity, and problem-solving \cite{gerecke2007challenges}. Figure \ref{fig:overview} shows the class structure from the author's recent instance of teaching an introductory human-robot interaction class. The author balanced theoretical components in class and independent activities like bi-weekly 75-minute lectures, daily readings, and quarterly homework assignments, with the experimental component of the semester-long group project. The group project featured multiple modes of assessment (based on the Universal Design for Learning \cite{rose2000universal}), including two forms of presentation (a midterm and a final) and a final report.
% \subsection{Theoretical}
% \label{subsec:theo}
% Fo
% \subsection{Experimental}
% \label{subsec:exp}

\section{The Case for Platform Standardization}
\label{sec:platform}
In addition to a standardized foundational theoretical basis for HRI courses across universities, students should have equal access to a standard robotic platform to apply this knowledge experimentally at the introductory level. To achieve this, HRI researchers need to either agree on a lower-cost commercially available platform that would be reasonable for universities to purchase for an introductory course (for example, the NAO from Softbank Robotics \cite{NAO} or Stretch from Hello Robot \cite{HelloRobot}, perhaps with the collaboration of a company), or come together to develop a multi-functional ultra-low cost robot for HRI educational purposes, similar to how the University of Michigan developed the MBot ecosystem \cite{MBot}. \textit{All universities, regardless of size or funding, should be able to offer the same quality of hands-on human-robot interaction education.}

% \textit{Students who attend better-funded universities should not receive a higher quality human-robot interaction education.} \looseness-1

\vspace{-0.25cm}
\section{The Case for Relevant Key Papers}
\label{sec:papers}
As the field of HRI is still relatively new, especially when compared to more established fundamental courses like dynamics or machine learning, students should remain abreast of the latest advances in the field. However, for a foundational course to be solely based on reading recent papers without instruction or hands-on experimentation would be a mistake. Thus, finding a healthy balance of technical content, project work, and reading relevant key papers is essential. At the introductory level, it is safe to assume that many students have had no experience reading research papers. Additionally, many graduate students receive no formal training in effectively reading research papers, though it is a critical skill for any future HRI researcher. Thus, I highly recommend first assigning the paper \textit{``How to read a paper''} and including a short, in-class review to ensure all students feel comfortable with the process \cite{keshav2007read}. Next, I have identified a few seminal papers related to various aspects of Human-Robot Interaction that I propose should be covered at some point during the semester. I do not believe these papers are comprehensive or the only critical readings from the field.  However, I think they span different vital areas, are well-written and researched, and provide a solid foundation for the new researchers to build on. 
\vspace{-0.15cm}
\begin{itemize}
    \item \underline{How to design a study:} \textit{``A Primer for Conducting Experiments in Human-Robot Interaction''} \cite{hoffman2020primer}
    \item \underline{How to evaluate HRI:} \textit{``Common Metrics for Human-Robot Interaction.''} \cite{Steinfeld2006}
    \item \underline{How to calculate and report statistical power:} \textit{``Have I Got the Power? Analyzing and Reporting Statistical Power in HRI.''} \cite{Bartlett2022}
    \item \underline{How to consider ethical implications:} \textit{``A Code of Ethics for the Human-Robot Interaction Profession''} \cite{riek2014code}
 \end{itemize}
 \vspace{-0.15cm}
% To learn how to design a research study, I recommend \textit{``A Primer for Conducting Experiments in Human-Robot Interaction''} by Hoffman and Zhao. To teach students the different ways to evaluate an interaction, I recommend Steinfeld et al.'s \textit{``Common Metrics for Human-Robot Interaction.''} It is also important that students understand the importance of conducting proper statistical analyses on their work, thus I recommend Bartlett et al.'s \textit{``Have I Got the Power? Analyzing and Reporting Statistical Power in HRI.''} Finally, ethical considerations should always be taken into account, so Riek and Howard's \textit{``A Code of Ethics for the Human-Robot Interaction Profession''} is an excellent resource to get students thinking about ethical implications of their work. 

After covering these articles, it can be up to the instructor's discretion whether they select additional papers for the students to read or whether the students are allowed to select the papers that pique their interests. The instructor should review and approve the student-selected papers to confirm that they are from recent (within the last ~3-5 years) relevant HRI conferences (e.g., HRI, ICSR, SORO, IROS, etc.). Individually or in a small group, students should present an overview of the paper in class, particularly looking at the manuscript with a critical lens for its strengths and weaknesses and presenting discussion questions for the class to debate. \looseness-1 %\textit{All three components: in-class instruction, hands-on project-based experimentation, and student-led discussions on relevant key papers are imperative for a comprehensive and successful foundational course in human-robot interaction.}

% \section{The Case Integrating Generative AI}
% \label{sec:genai}
% Integrating generative AI into the classroom environment bears a remarkable resemblance to the prior adoption of the Internet, and specifically Wikipedia, as educational tools. Much like the transformative impact of these resources, incorporating generative AI will enhance students' access to information and will encourage a dynamic learning experience. Similarly to how the Internet and Wikipedia provided quick and easy access to knowledge over the conventional World Book Encyclopedia, generative AI offers the ability to generate diverse and relevant content in real-time. These technologies empower students to explore beyond traditional textbooks, fostering critical thinking and research skills. While the adoption of the Internet and Wikipedia in the classroom faced initial skepticism, they ultimately became indispensable tools; similarly, embracing generative AI reflects a progressive approach to education, preparing students for the evolving landscape of information and technology.

\vspace{-0.3cm}
\section{Proposed Course Content}
\label{sec:content}
Based on my experience teaching a similar course this past semester, I propose an introductory HRI course should provide a sufficient high-level overview of all aspects of HRI, such that students should understand what options are available to them and can take further courses in their desired area of specific study later. Table \ref{tab:Course_content} shows a breakdown of the content I recommend including, with suggestions on the presentation order.

\begin{table}[b]
    \centering
    \begin{tabular}{|p{0.8cm}|p{7cm}|p{7cm}|}
        \hline
        \textbf{Week} & \textbf{Lecture Content} & \textbf{Project Milestone}\\
        \hline
        1-2 & Robot basics & \textbf{Introductions} (communicate interests, abilities, and skills)\\
        \hline
        2 & Design methods and protoyping & \textbf{Teaming} (form groups of 3 or 4 with common interests and complementary skills)\\
        \hline
        3-4 & Types of interactions (spatial, non-verbal, physical, verbal) & Project exploration (develop 3-5 ideas)\\
        \hline
        5 & Evaluating HRI systems & \textbf{Project proposal} (specify one idea and schedule for completion)\\
        \hline
        5-6 & User study design process & \textbf{Project outline} (identify project solution and narrative)\\
        \hline
        6 & Statistical analyses and reporting & \\
        \hline
        7 & Safety and ethics in HRI & \textbf{Infrastructure} (set up computational/mechanical tools/environment)\\
        \hline
        8 & Meetings and checkpoint presentations & \textbf{Benchmarking} (establish the minimum viable product of your HRI system)\\
        \hline
        9 & Emotion and Applicaitons in HRI & \textbf{Pilot study} (3-5 practice participants, adjust as needed)\\
        \hline
        10 & Scientific communication (making good figures and presentations) & \textbf{Iterate} (make necessary adjustments, \textbf{User study} (data collection, uniform protocol)\\
        \hline
        11 - 14 & Guest lectures and student-led paper presentations & \textbf{Initial results} (demonstrate progress), \textbf{Final results} (validate and support work)\\
        \hline
        15 & Final presentations & \textbf{Communication} (oral presentation, then 2-page extended abstract integrating feedback)\\
        \hline
    \end{tabular}
    \caption{Proposed course content for an introductory HRI course. The milestones for the semester-long hands-on HRI group project are tied to the lecture content presented in class.}
    \label{tab:Course_content}
    \vspace{-1cm}
\end{table}

\noindent \textbf{Robotics basics}: The interdisciplinary nature of HRI is one of the great strengths of the field. It also means an undergraduate HRI course likely will and should include students from multiple backgrounds, including some with no prior robotics experience. Thus, to ensure all students have a basic fundamental understanding of the elements of a robot, providing a single, ``crash-course style'' lesson on typical robotics sensors, actuators, and software on the first or second day of the semester will be critical.\\
% \noindent \textbf{Robot design and morphology}:\\
\noindent \textbf{Design methods and prototyping}: I believe it is important to share with students that they can follow several design processes to test their interactions. The three I recommend teaching are (1.) the Engineering Design Process, (2.) the User-Centered Design Process, and (3.) the Participatory Design Process. As part of this lesson, it is important to stress to the students the importance of engaging early and often with stakeholders (primary, secondary, and tertiary). Finally, the instructor should inform the students of all the prototyping resources available at the university (e.g., maker space, lab access).\\
\noindent \textbf{Types of interaction}: Presenting students with a high-level overview of all types of interactions possible with HRI (whether or not the instructor focuses their research in that area) is essentially providing students with a comprehensive and holistic understanding of the state-space of the field so they can determine which (if any) interactions they are most interested in researching further. Based on my experience this past semester, I propose presenting them in increasing order of complexity: \textit{spatial, non-verbal, physical, and verbal interaction} \\
\noindent \textbf{How to evaluate HRI systems}: Early-stage researchers need an introduction to the importance of using validated questionnaires, the benefits and drawbacks of different types of survey options (e.g., forced choice, graded-scale, visual analog scale, free-response), and the many quantitative data options available to evaluate user response to an interaction.\\
\noindent \textbf{The user study design process in HRI}: A thorough understanding of properly designing a user study for human-robot interaction is critical to ensure successful data collection. Too often, students are not trained in this essential component of research. As mentioned earlier, I recommend assigning the ``Primer'' \cite{hoffman2020primer}, and then using a \textit{flipped-classroom style} \cite{bishop2013flipped}, discuss the components of each step in the lecture the next day. Additionally, I have my students use this paper as a blueprint for conducting their user studies. After reviewing the paper together to ensure full understanding, the groups follow the steps outlined throughout the semester to conduct, analyze, and report their user studies.\\ 
%Therefore, I recommend educators of an introductory human-robot interaction course assign Hoffman and Zhao's ``A Primer for Conducting Experiments in Human-Robot Interaction'' for homework reading \cite{hoffman2020primer}, and then in a \textit{flipped-classroom style} \cite{bishop2013flipped}, discuss the components of each step in the lecture the next day. \\
\noindent \textbf{Statistical analyses and reporting statistics}: It is also important that students first entering the field of human-robot interaction have a basic understanding of how to analyze, interpret, and present data. Thus, I recommend having one unit presenting common statistical analyses used, how to apply them (students may or may not have taken an introductory statistics course yet), and how to report these statistical findings in a research paper for the community to understand the implications of their findings (i.e., important to not only report p-values but effect sizes as well). Some particularly important topics I recommend instructors cover are how to conduct a power analysis to identify how many participants researchers need for an experiment (educators may want to teach students to use G*Power \cite{GPower}), what the meaning of a significance level is, when and how to conduct a t-test, an ANOVA test, and posthoc analyses, at a minimum. \\
\noindent \textbf{Safety and ethical considerations in HRI}: While conducting a project with a user-study during a semester does not require ethical approval while it is in the context of a course because its purpose is for education, students must understand the process and important of applying for ethical approval for human-robot interaction user studies. Therefore, I recommend spending at the very least one class explaining the history of why ethical boards are necessary (e.g., Nazi experiments, Tuskegee Syphilis study), the evolution of the ethical regulations (e.g., Nuremberg Code, Declaration of Helsinki, Belmont Report, The Common Rule, the IRB), and the components a researcher must compile to submit to the regulating ethical board for approval to conduct a user study with human participants. If time is available, one way to make ethics more fun for students would be to turn the lesson into a debate, similar to the ICRA Robotics Debates \cite{ICRADebates}. In this case, the students could be separated into several topics (each with a team of students ``pro'' and ``against''), with the educator serving as the moderator. The educator can develop several controversial topics to have the students debate. For this to work effectively, I recommend giving the students at least one week to prepare. Example topics could include: (1) robots should replace all dull, dirty, and dangerous tasks currently performed by humans, (2) robots should be designed with the capacity to deceive humans to achieve certain goals, and (3) robots deserve equal rights as humans. The topics must strongly pick a side of a controversial topic so that the students can debate it. \\
\noindent \textbf{Emotions in robotics}: Emotions play a critical role in HRI and, thus, I believe, require an entire lecture. This lecture should explain the role of emotions in interactions and describe emotions, mood, and affect. As emotion classification is still a hotly contested issue, I propose presenting students with several different models of emotion (e.g., OCC model, Circumplex [valence-arousal], PANAS, PAD [pleasure, arousal, dominance]). Finally, it is essential to present to students the current challenges and limitations in affective HRI.\\
\noindent \textbf{Applications of HRI}: At the end of the semester, I recommend sharing several industries where HRI is applicable. Providing concrete, non-academic examples of HRI can help undergraduate students conceptualize where pursuing the field might take them. These applications can motivate student interest and broaden their understanding of HRI.\\
% \noindent \textbf{Future challenges for HRI}: For a more active-learning environment, educators can incorporate tools like Slido \cite{Slido}, so students can brainstorm together what they believe will be some of the biggest challenges for the field of HRI moving forward, and the instructor can help facilitate the discussion.\\
\noindent \textbf{Scientific communication}: Students are rarely trained in effective scientific communication, but this skill is critical to HRI. Not only for presenting research at scientific conferences but considering the significant media interest in HRI research, training students early on how to effectively and honestly represent their work to the public is important.\\
\noindent \textbf{Guest-lectures}: Bringing HRI researchers into the classroom provides students with a deeper dive into different areas of HRI and showcases what long-term research projects can yield. Guest lectures can make abstract concepts more concrete.\looseness-1\\
\noindent \textbf{Student-led paper presentations}: Following the guidelines outlined in Section~\ref{sec:papers} for reading and selecting scientific papers, students in the course should individually or in a small group present papers to the class to practice their scientific communication skills learned in an earlier lesson. To facilitate a more meaningful in-class discussion, educators can provide the other students with the paper being discussed beforehand to familiarize themselves with the content.\\ 
\noindent \textbf{Project milestones}: From the first lecture, students should be prepared and coached throughout the semester to complete a semester-long HRI team-based project. The instructor must set up interim checkpoints to keep students on task to complete the project within the semester. Teams should submit three ideas to the instructor and then meet with the instructor to discuss the feasibility of each idea, given the time frame and the teams' capabilities. Because user studies are integral to much of HRI research (though not all), I recommend that these semester projects feature small-scale user studies so students can learn that process. To help with participant recruitment, I suggest all students in the class serve as pilot participants for one other study and actual participants for two other projects. Finally, at the end of the semester, to practice the ``science communication'' skills, the project groups should present their work in a final presentation format, where the instructor invites other faculty from interdisciplinary departments (engineering and non-engineering [e.g., psychology]). Following feedback from their presentation during a short question and answer session, students should also submit a 2-page extended abstract describing their project and lessons learned.

\vspace{-0.4cm}
\section{Conclusion}
\label{sec:conc}
Standardizing an introductory course for HRI is a necessary endeavor that will require collaboration and agreement of experts from multiple areas of expertise within this interdisciplinary field. This manuscript proposed course content and advocated for the importance of hands-on experimental learning while pushing for platform equality, especially in these fundamental introductory courses. By establishing this standardized course, educators and institutions can ensure all students receive the same fundamental knowledge base that covers all facets of this complicated, exciting, and rapidly developing field. 
%\begin{acks}
%To Robert, for the bagels and explaining CMYK and color spaces.
%\end{acks}

\vspace{-0.4cm}
\bibliographystyle{ACM-Reference-Format}
\bibliography{refs}


\begin{thebibliography}{15}

%%% the Web address.

\ifx \showCODEN    \undefined \def \showCODEN     #1{\unskip}     \fi
\ifx \showDOI      \undefined \def \showDOI       #1{#1}\fi
\ifx \showISBNx    \undefined \def \showISBNx     #1{\unskip}     \fi
\ifx \showISBNxiii \undefined \def \showISBNxiii  #1{\unskip}     \fi
\ifx \showISSN     \undefined \def \showISSN      #1{\unskip}     \fi
\ifx \showLCCN     \undefined \def \showLCCN      #1{\unskip}     \fi
\ifx \shownote     \undefined \def \shownote      #1{#1}          \fi
\ifx \showarticletitle \undefined \def \showarticletitle #1{#1}   \fi
\ifx \showURL      \undefined \def \showURL       {\relax}        \fi
% invisible to TeX
\providecommand\bibfield[2]{#2}
\providecommand\bibinfo[2]{#2}
\providecommand\natexlab[1]{#1}
\providecommand\showeprint[2][]{arXiv:#2}

\bibitem[Bartlett et~al\mbox{.}(2022)]%
        {Bartlett2022}
\bibfield{author}{\bibinfo{person}{Madeleine~E. Bartlett},
  \bibinfo{person}{C.~E.~R. Edmunds}, \bibinfo{person}{Tony Belpaeme}, {and}
  \bibinfo{person}{Serge Thill}.} \bibinfo{year}{2022}\natexlab{}.
\newblock \showarticletitle{Have I Got the Power? Analysing and Reporting
  Statistical Power in HRI}.
\newblock \bibinfo{journal}{\emph{ACM Transactions on Human-Robot Interaction}}
  \bibinfo{volume}{11}, \bibinfo{number}{2}, Article \bibinfo{articleno}{16}
  (\bibinfo{date}{feb} \bibinfo{year}{2022}), \bibinfo{numpages}{16}~pages.
\newblock
\urldef\tempurl%
\url{https://doi.org/10.1145/3495246}
\showDOI{\tempurl}


\bibitem[Bishop and Verleger(2013)]%
        {bishop2013flipped}
\bibfield{author}{\bibinfo{person}{Jacob Bishop} {and}
  \bibinfo{person}{Matthew~A Verleger}.} \bibinfo{year}{2013}\natexlab{}.
\newblock \showarticletitle{The flipped classroom: A survey of the research}.
  In \bibinfo{booktitle}{\emph{ASEE Annual Conference \& Exposition}}.
  \bibinfo{publisher}{ASEE Conferences}, \bibinfo{address}{Atlanta, Georgia},
  \bibinfo{pages}{23--1200}.
\newblock


\bibitem[Buchner et~al\mbox{.}(2024)]%
        {GPower}
\bibfield{author}{\bibinfo{person}{Axel Buchner}, \bibinfo{person}{Edgar
  Erdfelder}, \bibinfo{person}{Franz Faul}, {and} \bibinfo{person}{Albert-Georg
  Lang}.} \bibinfo{year}{2024}\natexlab{}.
\newblock \bibinfo{title}{G*Power: Statistical Power Analyses for Mac and
  Windows}.
\newblock \bibinfo{howpublished}{Online}.
\newblock
\urldef\tempurl%
\url{https://www.psychologie.hhu.de/arbeitsgruppen/allgemeine-psychologie-und-arbeitspsychologie/gpower}
\showURL{%
Retrieved Jan, 2024 from \tempurl}


\bibitem[Gerecke and Wagner(2007)]%
        {gerecke2007challenges}
\bibfield{author}{\bibinfo{person}{Uwe Gerecke} {and} \bibinfo{person}{Bernardo
  Wagner}.} \bibinfo{year}{2007}\natexlab{}.
\newblock \showarticletitle{The challenges and benefits of using robots in
  higher education}.
\newblock \bibinfo{journal}{\emph{Intelligent Automation \& Soft Computing}}
  \bibinfo{volume}{13}, \bibinfo{number}{1} (\bibinfo{year}{2007}),
  \bibinfo{pages}{29--43}.
\newblock


\bibitem[Hoffman and Zhao(2020)]%
        {hoffman2020primer}
\bibfield{author}{\bibinfo{person}{Guy Hoffman} {and} \bibinfo{person}{Xuan
  Zhao}.} \bibinfo{year}{2020}\natexlab{}.
\newblock \showarticletitle{A primer for conducting experiments in human--robot
  interaction}.
\newblock \bibinfo{journal}{\emph{ACM Transactions on Human-Robot Interaction
  (THRI)}} \bibinfo{volume}{10}, \bibinfo{number}{1} (\bibinfo{year}{2020}),
  \bibinfo{pages}{1--31}.
\newblock


\bibitem[Keshav(2007)]%
        {keshav2007read}
\bibfield{author}{\bibinfo{person}{Srinivasan Keshav}.}
  \bibinfo{year}{2007}\natexlab{}.
\newblock \showarticletitle{How to read a paper}.
\newblock \bibinfo{journal}{\emph{ACM SIGCOMM Computer Communication Review}}
  \bibinfo{volume}{37}, \bibinfo{number}{3} (\bibinfo{year}{2007}),
  \bibinfo{pages}{83--84}.
\newblock


\bibitem[Pashler et~al\mbox{.}(2008)]%
        {pashler2008learning}
\bibfield{author}{\bibinfo{person}{Harold Pashler}, \bibinfo{person}{Mark
  McDaniel}, \bibinfo{person}{Doug Rohrer}, {and} \bibinfo{person}{Robert
  Bjork}.} \bibinfo{year}{2008}\natexlab{}.
\newblock \showarticletitle{Learning styles: Concepts and evidence}.
\newblock \bibinfo{journal}{\emph{Psychological science in the public
  interest}} \bibinfo{volume}{9}, \bibinfo{number}{3} (\bibinfo{year}{2008}),
  \bibinfo{pages}{105--119}.
\newblock


\bibitem[Riek and Howard(2014)]%
        {riek2014code}
\bibfield{author}{\bibinfo{person}{Laurel Riek} {and} \bibinfo{person}{Don
  Howard}.} \bibinfo{year}{2014}\natexlab{}.
\newblock \showarticletitle{A code of ethics for the human-robot interaction
  profession}.
\newblock \bibinfo{journal}{\emph{Proceedings of we robot}}
  (\bibinfo{year}{2014}).
\newblock


\bibitem[Robot(2024)]%
        {HelloRobot}
\bibfield{author}{\bibinfo{person}{Hello Robot}.}
  \bibinfo{year}{2024}\natexlab{}.
\newblock \bibinfo{title}{Stretch Research Edition}.
\newblock \bibinfo{howpublished}{Online}.
\newblock
\urldef\tempurl%
\url{https://hello-robot.com/product}
\showURL{%
Retrieved Jan, 2024 from \tempurl}


\bibitem[Robotics(2023)]%
        {MBot}
\bibfield{author}{\bibinfo{person}{UM Robotics}.}
  \bibinfo{year}{2023}\natexlab{}.
\newblock \bibinfo{title}{The MBot Ecosystem}.
\newblock \bibinfo{howpublished}{Online}.
\newblock
\urldef\tempurl%
\url{https://mbot.robotics.umich.edu}
\showURL{%
Retrieved Jan, 2024 from \tempurl}


\bibitem[RobotLab(2024)]%
        {NAO}
\bibfield{author}{\bibinfo{person}{RobotLab}.} \bibinfo{year}{2024}\natexlab{}.
\newblock \bibinfo{title}{NAO Robot Power V6 Standard Edition for Research}.
\newblock \bibinfo{howpublished}{Online}.
\newblock
\urldef\tempurl%
\url{https://www.robotlab.com/higher-ed-robots/store/nao-power-v6-standard-edition}
\showURL{%
Retrieved Jan, 2024 from \tempurl}


\bibitem[Rose(2000)]%
        {rose2000universal}
\bibfield{author}{\bibinfo{person}{David Rose}.}
  \bibinfo{year}{2000}\natexlab{}.
\newblock \showarticletitle{Universal design for learning}.
\newblock \bibinfo{journal}{\emph{Journal of Special Education Technology}}
  \bibinfo{volume}{15}, \bibinfo{number}{4} (\bibinfo{year}{2000}),
  \bibinfo{pages}{47--51}.
\newblock


\bibitem[Steinfeld et~al\mbox{.}(2006)]%
        {Steinfeld2006}
\bibfield{author}{\bibinfo{person}{Aaron Steinfeld}, \bibinfo{person}{Terrence
  Fong}, \bibinfo{person}{David Kaber}, \bibinfo{person}{Michael Lewis},
  \bibinfo{person}{Jean Scholtz}, \bibinfo{person}{Alan Schultz}, {and}
  \bibinfo{person}{Michael Goodrich}.} \bibinfo{year}{2006}\natexlab{}.
\newblock \showarticletitle{Common metrics for human-robot interaction}. In
  \bibinfo{booktitle}{\emph{ACM SIGCHI/SIGART International Conference on
  Human-Robot Interaction}} (Salt Lake City, Utah, USA)
  \emph{(\bibinfo{series}{HRI '06})}. \bibinfo{publisher}{Association for
  Computing Machinery}, \bibinfo{address}{New York, NY, USA},
  \bibinfo{pages}{33–40}.
\newblock
\showISBNx{1595932941}
\urldef\tempurl%
\url{https://doi.org/10.1145/1121241.1121249}
\showDOI{\tempurl}


\bibitem[Stice(1987)]%
        {stice1987using}
\bibfield{author}{\bibinfo{person}{James~E Stice}.}
  \bibinfo{year}{1987}\natexlab{}.
\newblock \showarticletitle{Using Kolb's Learning Cycle to Improve Student
  Learning.}
\newblock \bibinfo{journal}{\emph{Engineering education}} \bibinfo{volume}{77},
  \bibinfo{number}{5} (\bibinfo{year}{1987}), \bibinfo{pages}{291--96}.
\newblock


\bibitem[Team(2022)]%
        {ICRADebates}
\bibfield{author}{\bibinfo{person}{The Robotics~Debates Team}.}
  \bibinfo{year}{2022}\natexlab{}.
\newblock \bibinfo{title}{Debates on the Future of Robotics Research}.
\newblock \bibinfo{howpublished}{Online}.
\newblock
\urldef\tempurl%
\url{https://www.roboticsdebates.org}
\showURL{%
Retrieved Jan, 2024 from \tempurl}


\end{thebibliography}

\end{document}